\documentclass[sigconf]{acmart}

\usepackage{booktabs} 
\usepackage{subfigure}
\usepackage{multirow}
\usepackage{url}
\usepackage{enumerate}
\usepackage{colortbl}
\usepackage{float}
\usepackage{graphicx}
\usepackage{amssymb}
\usepackage{amsmath}
\usepackage{array}
\usepackage{algorithm} 
\usepackage{algorithmicx} 
\usepackage{algpseudocode}  
\usepackage{caption}
\usepackage{bm}

\acmDOI{10.475/123_4}

\acmISBN{123-4567-24-567/08/06}

\acmConference[]{}{}{} 
\acmYear{2018}
\copyrightyear{2018}

\acmPrice{15.00}

\begin{document}
\title{DeepMove: Learning Place Representations through Large Scale Movement Data}

\author{Yang Zhou}
\affiliation{%
  \institution{Department of Computer Science and Engineering}
  \streetaddress{University of North Texas}
  \city{Denton}
  \state{Texas}
  \postcode{76203}
}
\email{yangzhou2@my.unt.edu}

\author{Yan Huang}
\affiliation{%
  \institution{Department of Computer Science and Engineering}
  \streetaddress{University of North Texas}
  \city{Denton}
  \state{Texas}
  \postcode{76203}
}
\email{yan.huang@unt.edu}


\begin{abstract}
Understanding and reasoning about places and their relationships are critical for many applications.  Places are traditionally curated by a small group of people as place gazetteers and are represented by an ID with spatial extent, category, and other descriptions. However, a place context is described to a large extent by movements made from/to other places. Places are linked and related to each other by these movements. This important context is missing from the traditional representation. 

We present DeepMove, a novel approach for learning latent representations of places.  DeepMove advances the current deep learning based place representations by directly model movements between places. We demonstrate DeepMove's latent representations on place categorization and clustering tasks on large place and movement datasets with respect to important parameters. Our results show that DeepMove outperforms state-of-the-art baselines. DeepMove's representations can provide up to 15\% higher than competing methods in matching rate of place category and result in up to 39\% higher silhouette coefficient value for place clusters.

DeepMove is spatial and temporal context aware. It is scalable. It outperforms competing models using much smaller training dataset (a month or 1/12 of data). These qualities make it suitable for a broad class of real-world applications.
\end{abstract}

%
%


\maketitle

\section{Introduction}
People move from one place to another constantly. Places are linked and related to each other by these movements. A place can be a point of interest such as shopping mall, gym, bank, or historical landmark. A place can also be a region such as a residential neighborhood.
Places are traditionally represented by an ID with longitude, latitude, category, and other descriptions in current systems of location recommendation systems, city function modeling, transportation planning, bike flower prediction, place classification, and trajectory analysis. However, a place context is described to large extent by incoming and outgoing movements made. This context is missing from the traditional representation. 

In this paper we introduce deep learning (unsupervised learning) techniques, which have been shown impressive success in natural language processing \cite{rumelhart_learning_1986,bengio_neural_2003,collobert_unified_2008,reisinger_multi-prototype_2010,mikolov_distributed_2013, mikolov_efficient_2013, mikolov_linguistic_2013,pennington_glove:_2014,levy_improving_2015}, information networks \cite{dong_metapath2vec:_2017,tang_line:_2015,sun2009rankclus}, and graph analytics \cite{Xie:2016:LGP:2983323.2983711,yang_revisiting_2016,perozzi_deepwalk:_2014,cao_grarep:_2015,nguyen2013lightweight}. In natural language processing tasks, each word is embedded in a vector which can be seen as representing the distribution of the context in which a word appears. The vector encodes many linguistic regularities and patterns. In information networks and graph analytics, embedding has significantly improved results in multi-label classification and graph visualization. 

With the availability of large movement datasets collected by personal location devices and location services (Waze, Uber, Didi, Google Map), it is possible to train more complex machine learning models to drive fine grained spatial and temporal context of places. Large-scale movement data with better learning algorithms provide an opportunity to innovate traditional place representation models and allow more sophisticated place reasoning.

\begin{figure*}
\begin{minipage}[t]{0.3\linewidth}
\centering
\includegraphics[width=2in, height=2.0in]{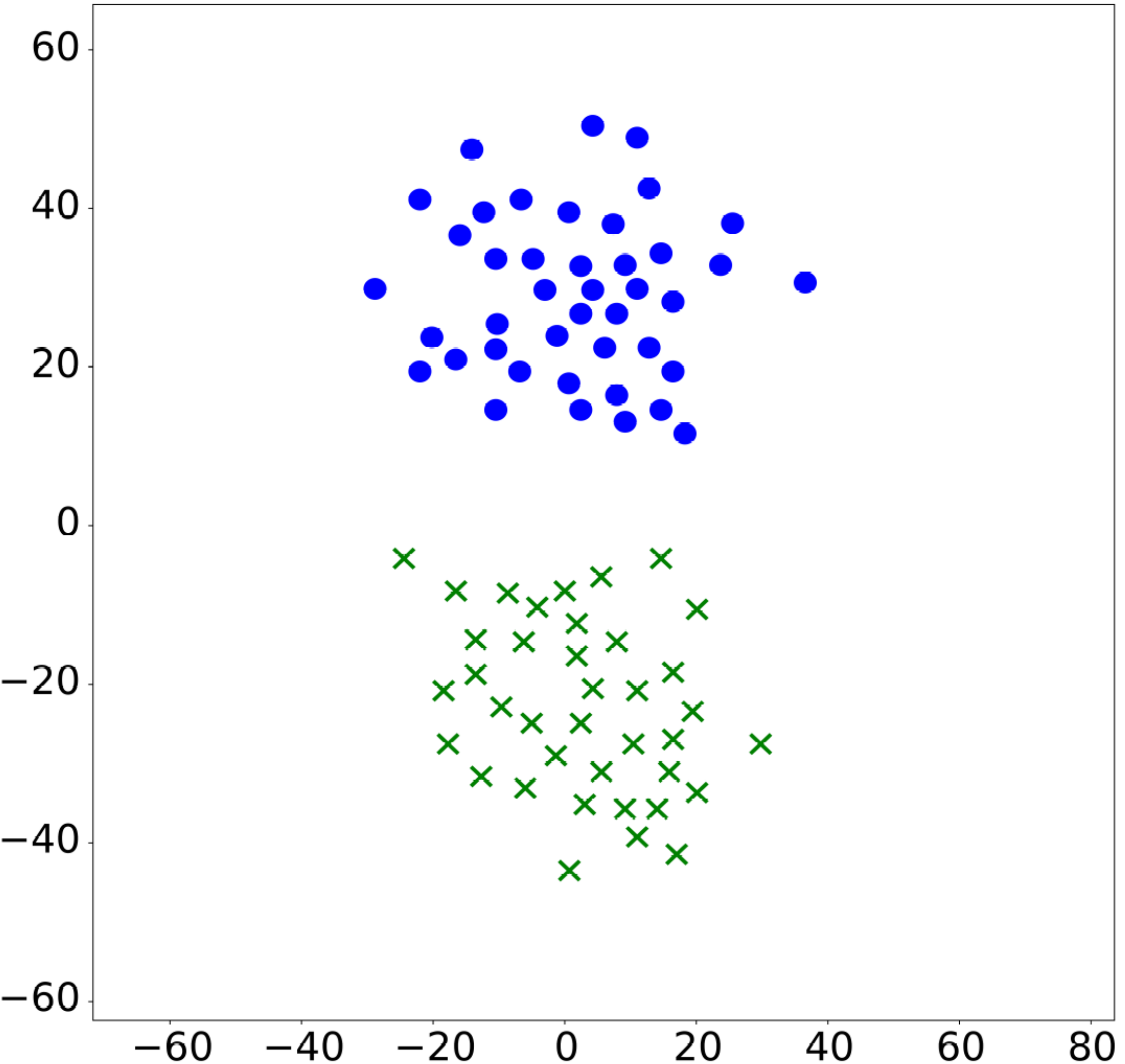}
\end{minipage}%
\begin{minipage}[t]{0.3\linewidth}
\centering
\includegraphics[width=2in, height=2.0in]{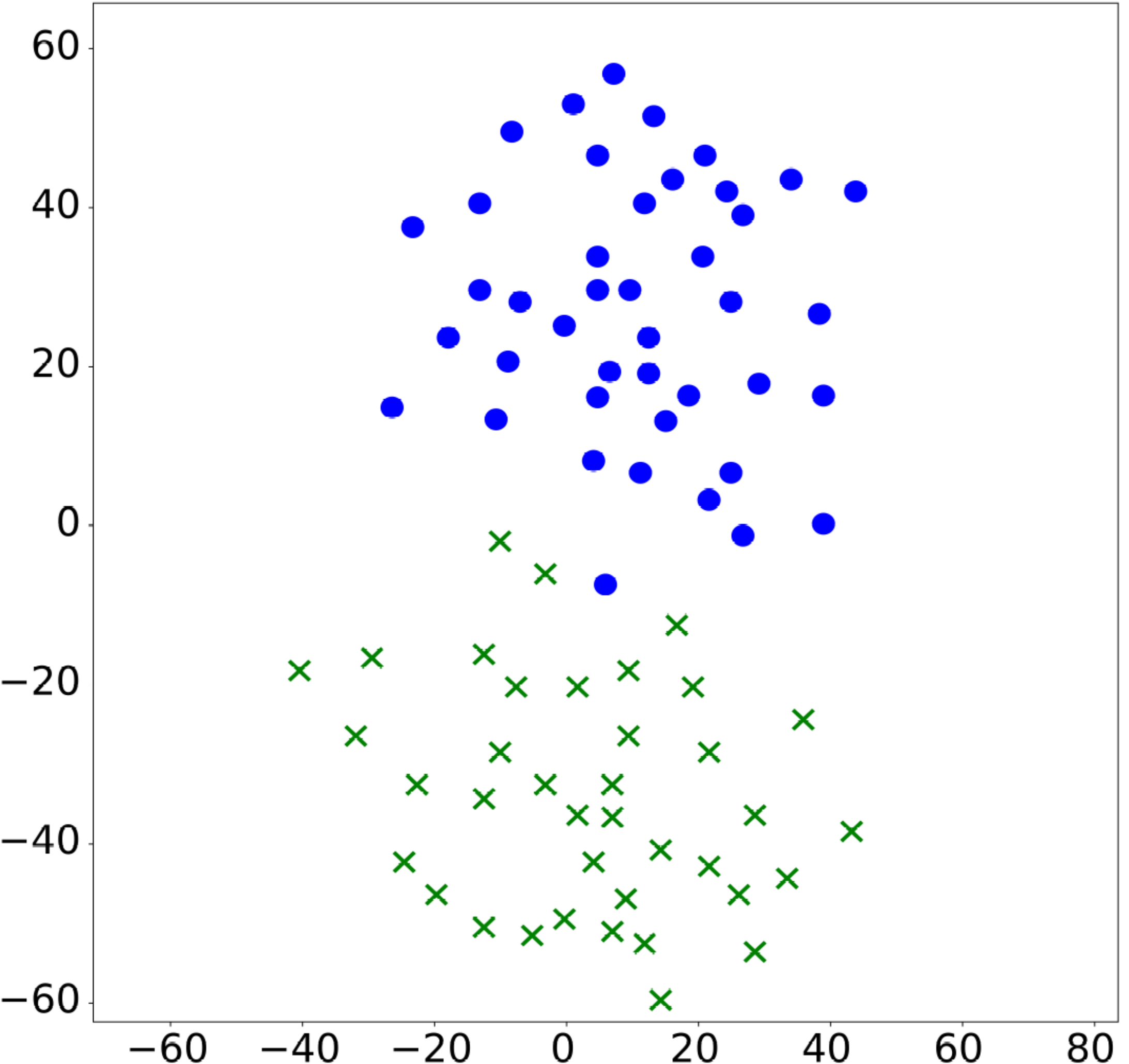}
\end{minipage}
\begin{minipage}[t]{0.3\linewidth}
\centering
\includegraphics[width=2in,height=2.0in]{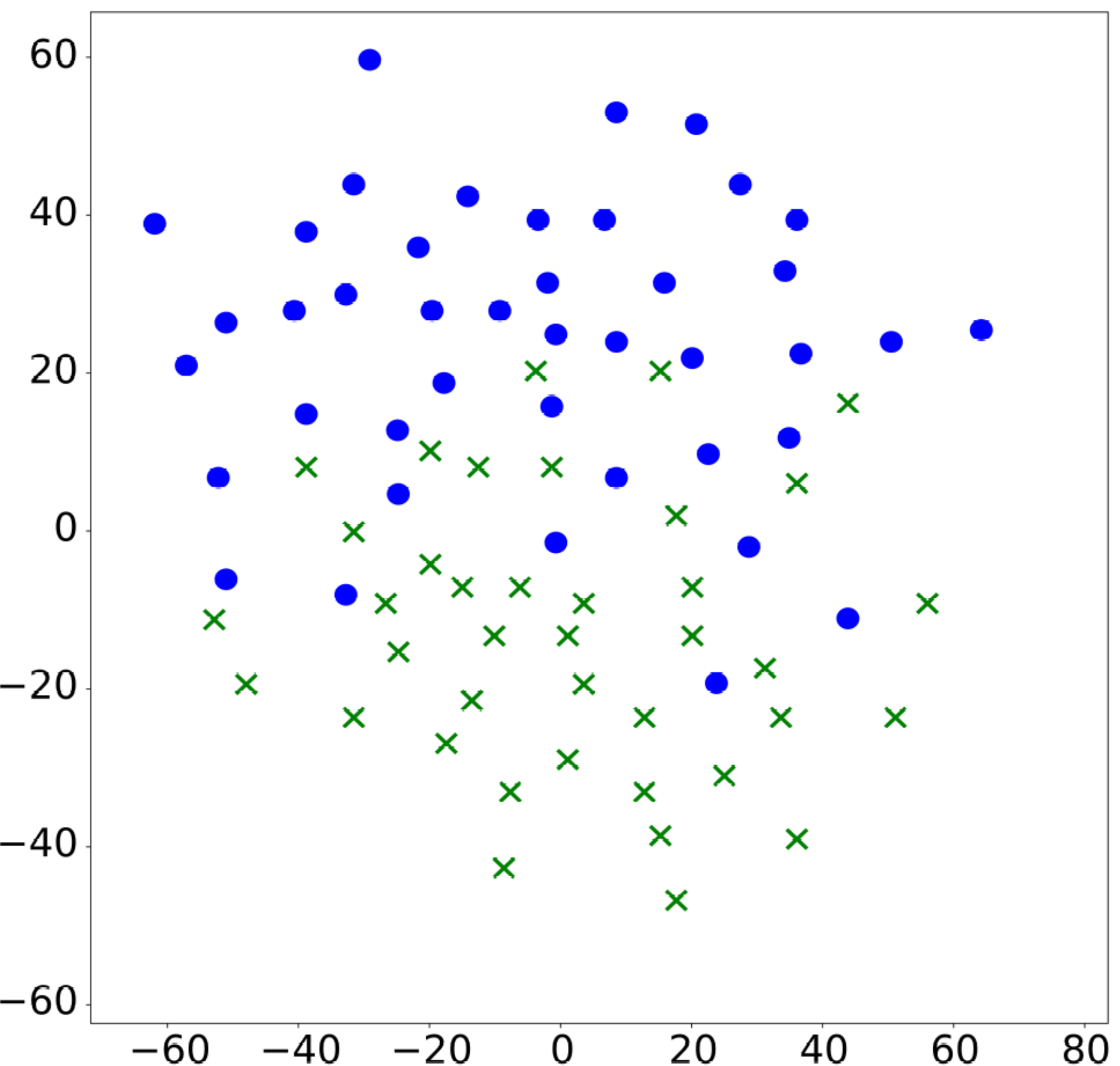}
\end{minipage}
\caption{Our proposed DeepMove (left two figures) learns latent space representations that separate Residential and
Recreational places substantially better than the state-of-art ITDL model \cite{yan2017itdl} (right figure) on New York City datasets \cite{newyork}. Here, standard dimension reduction method t-SNE \cite{maaten2008visualizing} is used to reduce the 180 dimensional embedding to 2 dimensional space. The learned representations of DeepMove encode spatial and temporal movement context of places so they can be easily exploited by many learning methods.}
\label{fig:tsne}
\end{figure*}

In this paper, we propose DeepMove to characterize places and their relationships by integrating large scale movement data. Specifically, DeepMove uses movement information to learn the latent representations of places, which preserves spatial and temporal context. Our contributions are as follows:
\begin{itemize}
\item We propose DeepMove, a novel approach for learning latent representations of places. These latent representations encode place relations in a continuous vector space. DeepMove generalizes recent advancements in language modeling and unsupervised feature learning (or deep learning) from sequences of words and graphs to places. DeepMove advances the current deep learning based place representations by directly modeling movements between places.

\item We demonstrate DeepMove's latent representations on place categorization and clustering tasks on large place and movement datasets with respect to four important parameters: dimension  of latent representation, time interval, number of epochs, and learning data sizes. Our results show that DeepMove outperforms state-of-the-art baselines. DeepMove's representations can provide up to 15\% higher than competing methods in matching rate of place category and result in up to 39\% higher silhouette coefficient value for place clusters.

\item DeepMove is spatial and temporal context aware. It is scalable. It outperforms competing models using much smaller training dataset (a month or 1/12 of data). These qualities make it suitable for a broad class of real-world applications such as place classification, geographical information retrieval, location-based recommendation, region function modeling, and urban space understanding.

\end{itemize}

The results of applying DeepMove and the current state-of-the-art method to the New York City point of interest and taxi trip datasets \cite{newyork} are shown in Figure \ref{fig:tsne}. Our proposed OD model and Trip model are shown in the left. The competing place embedding model \cite{yan2017itdl} is shown on the right.  Results show Residential and Recreational places in 2 latent dimensions after dimension reduction using t-Distributed Stochastic Neighbor Embedding (t-SNE) \cite{maaten2008visualizing} technique. Our models embed residential (blue dots) and recreational (green crosses) places into two cleanly-separated clusters. 

The rest of the paper is organized as follows.  In Section 2, we discuss preliminaries of place embedding and the properties of place data that allow the proposed representations to be successful. In Section 3, we give an introduction of Skip-gram model, and propose our trip model and OD model. In Section 4, we outline our experiments and presents our results. In Section 5, we summarize related work. Finally, in Section 6, we close with our conclusion.

\section{Preliminaries}
In this section, we introduce place embedding and the unique characteristics of places and their relationships. Then, we describe the connection between our work and natural language model by analyzing the data distribution.
\subsection{Place Embedding}
Places form the nodes of a network and movements happen between places. In information network or graph embedding, two nodes are considered similar and embedded in similar vectors if information diffuses between them or one reaches the other easily, e.g., through the random walk \cite{perozzi_deepwalk:_2014}. Place network has its distinct spatial-temporal characteristics. Movements follow a gravity model that dominates the macroscopic relationships between places. It is well known that interactions between two places decline with increasing (distance, time, and cost) and are also determined by variables including demographic gravitation and categories of the places. While a general network embedding can construct a network of places using distances or number of movements between them as weights, this embedding will result in a latent representation that encodes place connectivies. However,  when one moves from one place to another, the two places are usually different, e.g., from home to work. However, places of similar functional purposes typically have the similar origin or destination distributions. For example, a residential house usually generates trips towards work, entertainment, and shopping. In this paper, we model movement distributions instead of connectivities.

\subsection{Power Laws}
\begin{figure}[ht]
        \centering
        \includegraphics[width=0.9\linewidth]{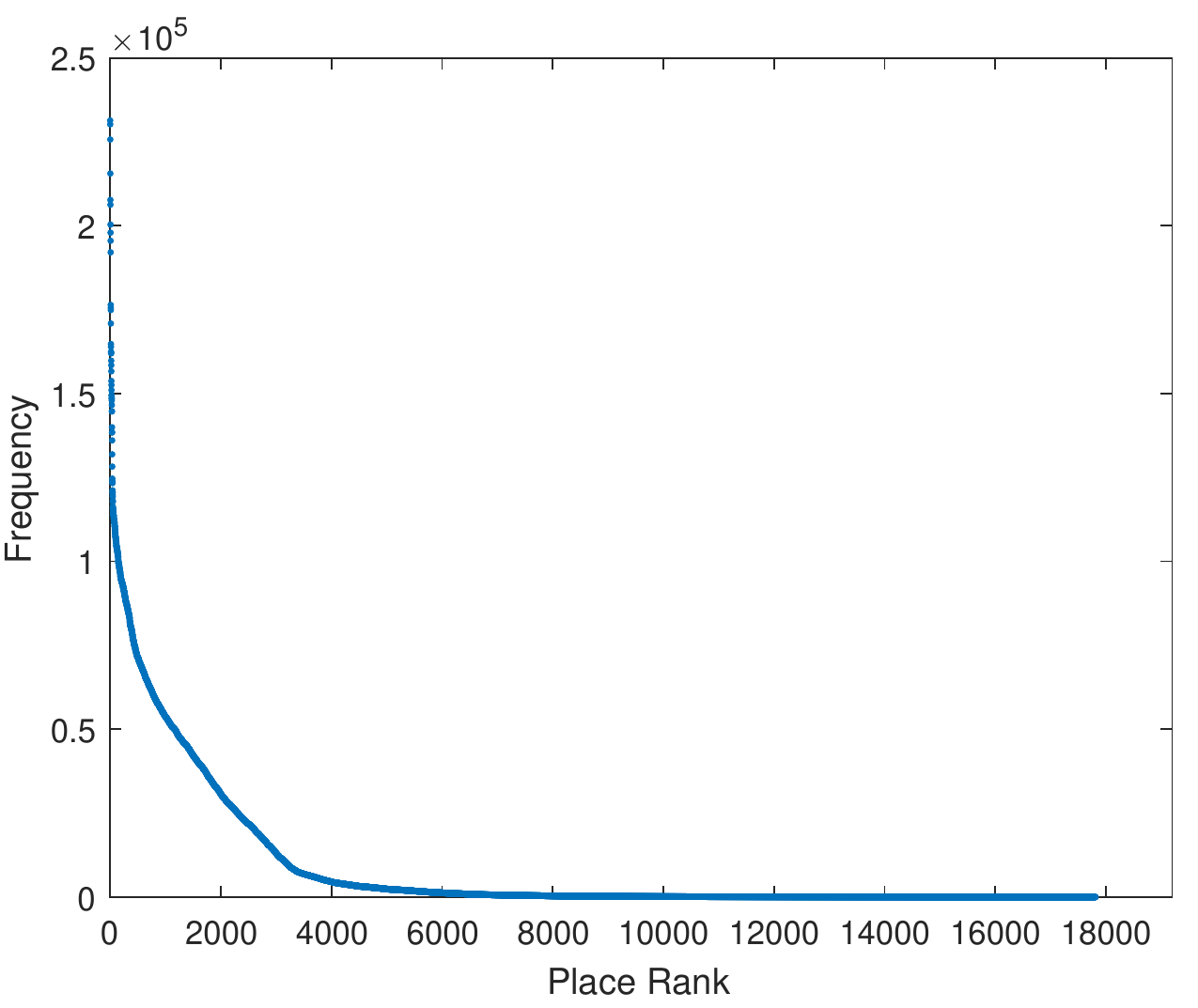}
        \caption{Place Rank-Frequency Plot}
        \label{fig:power_law}
\end{figure}

Words frequency in natural language follows a power law distribution. Previous work shows that the latent representation method can effectively model the sequential semantic relationships among words based on the characteristic of this word distribution. We generate Figure \ref{fig:power_law} using trip origins in the
New York City Taxi \& Limousine Commission \cite{newyork} yellow and green taxi trip data between 2009-2017. From Figure \ref{fig:power_law} we can see that, the distribution of place frequency also follows   power law distribution as the word frequency distribution, exhibiting the long tail phenomenon. Moreover, we also find a linear trend between log(place rank) and log(frequency) by using linear regression. The R-square of the linear regression is 0.852, and the p-value is less than 0.001, which means the model can fit the data very well. All these statistics show that the trip origins follow the power law distribution, which allows us to use deep learning models the place context. 

\section{Methods}
The recently proposed spatial context model considers a place nearby as the context of the other \cite{yan2017itdl}. This model will result in embedding two places similarly when they are in spatial proximity. This embedding captures spatial closeness but fails in encoding movement information.  So, in this section, we first introduce Skip-gram model, which is the model our work is based on. After that, we propose two models that incorporate movements: a Trip Model and an Origin-Destination (OD) Model.

\subsection{SkipGram}
The training purpose of Skip-gram model \cite{mikolov2013distributed} is to predict a word's context given the word itself. The model tries to maximize the probability of co-occurrence of words within a window in a sentence. Formally, we define $ V $ as the vocabulary, and define $W$ as the training set, which contains a sequence of training words $ w_{1}, w_{2}, w_{3},..., w_{T} $, the objective of the Skip-gram model is to maximize the average log probability:  

\begin{equation}
Pr(\{ w_{n-c},..., w_{n+c}\} \backslash w_{n}| w_{n}) = \frac{1}{T}\sum_{n=1}^{T}\sum_{\begin{subarray}
\ -c\leq m\leq c \\ \ \ \ \ m\neq 0\end{subarray}}^{} log\ p(w_{n+m}|w_{n}) 
\end{equation}
where $ c $ is the window size, $ T = |W| $ is the number of words in the training set, and the center word is $ w_{n} $, and $ w_{n+m}\ (-c\leq m\leq c)$ is context word. We use softmax function to define the probability $ p(w_{n+m}|w_{n}) $ from Skip-gram formulation, which is shown in equation \ref{eq:softmax}:

\begin{equation}
\label{eq:softmax}
p(w_{n+m}|w_{n}) = \frac{exp(u_{w_{n+m}}^{\intercal}v_{w_{n}})}{\sum_{k=1}^{T}exp(u_{w_{k}}^{\intercal}v_{w_{n}})}
\end{equation}
we introduce two mapping functions $ \Phi: w \in V \rightarrow \mathbb{R}^{T\times d} $ and $ \Phi': w \in U \rightarrow \mathbb{R}^{T\times d} $, $ d $ is vector's dimension. The first mapping function $\Phi$ maps each center word $w_{n}$ into the vector representation $v_{w_{n}}$. The second mapping function $\Phi'$ maps each context word $w_{n+m}$ into the vector representation $u_{w_{n+m}}$. So in equation \ref{eq:softmax}, $ v_{w_{n}} $ and $ u_{w_{n+m}} $ are the center and context vector representations of word $ w_{n} $ and word $ w_{n+m}$. We define $ C_{w_{n}} $ as the context set of word $ w_{n} $, and all the $ C_{w_{n}} $ where $ (1\leq n\leq T) $ forms the set $ C $, which we called total context set. 

Algorithm \ref{skigram} shows the pseudo code of Skip-gram. For each word $ w$, it iterates over all possible collocations that appear in $ C_{w} $ (Lines 1). We then map each word $ w $ and its context word $ w_{m} $ to their current representation vectors $\Phi(w), \Phi'(w_{m})\in \mathbb{R}^{d}$. By given the representation of $ w$ and $ w_{m}$, we would like to maximize the probability (Line 2). Skip-gram model needs huge computational resources, which is impractical for large datasets. To deal with this problem and speed up the training, we use the negative sampling technique \cite{goldberg2014word2vec} to modify the optimization objective, which causes each training sample to update only a small percentage of the model's weight, rather than all of them. 
 \begin{algorithm}[h]  
   \caption{SkipGram($ \Phi, w, C_{w} $)}  
   \begin{algorithmic}[1]  
       
       \State \textbf{for each} $ w_{m} \in C_{w}$ \textbf{do}
       \State \ \ \ \ \ \ \ \ $ J(\Phi)$ = -log Pr$(\Phi'(w_{m})|\Phi(w))$
       \State \ \ \ \ \ \ \ \ $ \Phi = \Phi$ - $ \alpha * \dfrac{\partial J}{\partial \Phi} $
       \State  \ \ \ \ \textbf{end for}
       \State \textbf{end for}
   \end{algorithmic}  
   \label{skigram}
 \end{algorithm}  
\subsection{Trip Model}
 Trips represent the movements from one place to another, which are important context when building place embedding. A trip model incorporates trip information when building the spatial context model. The key idea of the trip model is taking the destination as the context of the origin in a trip. For example, if there is a trip from restaurant to workplace, then we form a pair as $ (restaurant, workplace) $, and workplace is the context of the origin place restaurant. Let $P=\{p_1, p_2,\ldots, p_N\}$ be the set of places, and each place is different from the others. So $|N|$ is defined as the total number of different places. We use $ M $ to stand for trip set and each trip can be defined as $ Tr_{i} $, which is a tuple of $(o_i, d_i, ot_i, dt_i)$ where $ ( 1 \leqslant i \leqslant \lvert M\rvert) $. $ \lvert M\rvert $ is the total number of trips. For trips, it could have multiple trips with same origins or destinations. In the trip tuple, $o_i \in P$ is the origin place of the trip, $d_i \in P$ is the destination place of the trip, where $ot_i$/$dt_i$ represents the time leaving $o_i$/arriving $d_i$. The trip model is formalized by maximizing the average log probability of:

\begin{equation}
\label{eq:trip_model_log}
\frac{1}{M} \sum_{i=1}^M {\log p(d_i|o_i)}
\end{equation} 
The probability $p(d_i|o_i)$ is defined by using the softmax function:

\begin{equation}
\label{eq:trip_model_softmax}
p(d_i|o_i) = \frac{exp({u_{d_i}}^Tv_{o_i})}{\sum_{k=1}^{M}exp({u_{d_{k}}}^Tv_{o_i})}
\end{equation} 
where $u_{d_{k}}$ is the vector representation of a destination place $d_{k}$, $v_{o_{i}}$ and $u_{d_{i}}$ are the vector representation of origin place $o_{i}$ and destination place $d_{i}$.
\begin{algorithm}[h]  
   \caption{Trip Model ($ M, |N|, d, \gamma$)}  
   \begin{algorithmic}[1]  
   \renewcommand{\algorithmicrequire}{\textbf{Input:}} 
   \renewcommand{\algorithmicensure}{\textbf{Output:}}
       \Require trip set M    
       \renewcommand{\algorithmicrequire}{\textbf{}}   
       \Require \ \ \ \ \ \ total number of different places $|N|$
       \Require \ \ \ \ \ \ embedding size d
        \Require \ \ \ \ \ \ iteration number $\gamma$
       \Ensure matrix of place representations $ \Phi \in \mathbb{R}^{|N|\times d} $
       \State Initialization: Initialize $\Phi$ by using uniform distribution
       \State \textbf{for} $ i = 0 $ to $\gamma$ \textbf{do}
       \State  \ \ \ \ \textbf{for each} $ Tr_{i} \in M $ \textbf{do}
       \State  \ \ \ \ \ \ \ \ \ get $ o_{i} $, $ d_{i} $ from $ Tr_{i}$
       \State  \ \ \ \ \ \ \ \ \ SkipGram($ \Phi, o_{i}, d_{i} $)
       \State  \ \ \ \ \textbf{end for}
       \State \textbf{end for}
   \end{algorithmic}  
   \label{tripmodel}
 \end{algorithm}  

Line 2-7 in algorithm \ref{tripmodel} shows the core of our Trip model. The outer loop specifies the total number of iterations $ \gamma $, which also called the number of epochs. For each iteration, the entire data is passed through the model. We will discuss the performance with respect to the different number of epochs in the experiment part later. In the inner loop, we traverse all trips from the trip dataset. For each trip, we get the origin and destination POI and then pass it to Skip-gram model to update the representations (Line 5).   

\subsection{OD Model}
An OD model also uses the trip information to help build the spatial context model. However, different from the trip model, an OD Model takes the origin of another trip with the same destination as the context. For example, if there are two trips, the first one is from restaurant to workplace, the second one is from coffee store to the same workplace, then we get a pair, which is defined as (\textit{restaurant}, \textit{coffee store}). In this pair, the origin is a restaurant, and a coffee store is the context. Time is important when analyzing movements. People will have different movements towards different time periods. For a simple instance, many people will go to work from home around 8-9 am, and return from the workplace to the home around 5-6 pm. So, it will cause two different movements: home to workplace and workplace to home, which are based on the different temporal information. Therefore, the temporal aspect of movements is a critical factor for place representations. Two places may have movements to similar destinations at the same time. With different time intervals, the movements of places may vary. In OD model, we incorporate temporal information when building the context model. There is a surprisingly elegant and straightforward way to model time. We propose to introduce time interval to the modeling and with the goal to maximize the average log probability of:

\begin{equation}
\label{eq:od_model_log}
\frac{1}{M} \sum_{i=1}^M \sum_{\begin{subarray}\ \ \ \
d_j=d_i, j \neq i \\ |dt_i-dt_j| \leq W_d\end{subarray}}{\log p(o_j|o_i)}
\end{equation} 
where $ d_j=d_i, j \neq i  $ means there are two different trips with the same destination place. $W_d$ is a time interval to constrain trips reaching destination, e.g. 1 hour. The probability $p(o_j|o_i)$ is defined by using the softmax function:

\begin{equation}
\label{eq:od_model_softmax}
p(o_j|o_i) = \frac{exp({u_{o_j}}^Tv_{o_i})}{\sum_{k=1}^{M}exp({u_{o_{k}}}^Tv_{o_i})}
\end{equation} 
where $u_{o_{k}}$, $u_{o_{j}}$, and $v_{o_{i}}$ are the vector representations of origin place $o_{k}$, $o_{j}$ and $o_{i}$. 

\begin{algorithm}[h]  
   \caption{OD Model ($ M, |N|, d, \gamma, W_{d}$)}  
   \begin{algorithmic}[1]  
   \renewcommand{\algorithmicrequire}{\textbf{Input:}} 
   \renewcommand{\algorithmicensure}{\textbf{Output:}}
       \Require trip set M    
       \renewcommand{\algorithmicrequire}{\textbf{}}     
       \Require \ \ \ \ \ \ total number of different places $|N|$
       \Require \ \ \ \ \ \ embedding size d
        \Require \ \ \ \ \ \ iteration number $\gamma$
        \Require \ \ \ \ \ \ time interval $w_{d}$
       \Ensure matrix of place representations $ \Phi \in \mathbb{R}^{|N|\times d} $
       \State Initialization: Initialize $\Phi$ by using uniform distribution
       \State \textbf{for} $ i = 0 $ to $\gamma$ \textbf{do}
       \State  \ \ \ \ \textbf{for each} $ Tr_{i} \in M $ \textbf{do}
       \State  \ \ \ \ \ \ \ \ \ initialize origin set $ O $
       \State  \ \ \ \ \ \ \ \ \ get $ d_{i} $, $o_{i}$, $d_{t_{i}}$ from $ Tr_{i}$
       \State  \ \ \ \ \ \ \ \ \ \textbf{for each} $ Tr_{j} \in M $ and $ Tr_{j} \neq Tr_{i} $ \textbf{do}
       \State  \ \ \ \ \ \ \ \ \ \ \ \ \ get $ o_{j} $, $d_{j}$, $d_{t_{j}}$ from $ Tr_{j}$
       \State  \ \ \ \ \ \ \ \ \ \ \ \ \ \textbf{if} $d_{j}=d_{i}$ and $ |d_{t_{i}}-d_{t_{j}}| \leq W_{d} $ \textbf{do}
       \State  \ \ \ \ \ \ \ \ \ \ \ \ \ \ \ \ \ append $ o_{j} $ in $ O $
       \State  \ \ \ \ \ \ \ \ \ \ \ \ \ \textbf{else} 
       \State  \ \ \ \ \ \ \ \ \ \ \ \ \ \ \ \ \ \textbf{pass}
       \State  \ \ \ \ \ \ \ \ \ \textbf{if} $ O $ is not empty \textbf{do}
       \State  \ \ \ \ \ \ \ \ \ \ \ \ \ SkipGram($ \Phi, o_{i}, O $)
       \State  \ \ \ \ \ \ \ \ \ \textbf{else}
       \State  \ \ \ \ \ \ \ \ \ \ \ \ \ \textbf{pass}
       \State  \ \ \ \ \ \ \ \ \ \textbf{ end for} 
       \State  \ \ \ \ \textbf{end for}
       \State \textbf{end for}
   \end{algorithmic}  
   \label{odmodel}
 \end{algorithm}  

Algorithm \ref{odmodel} gives the description of our OD model. Same with trip model, in the first for loop, we also specify an iteration number to run our model (Line 2). In the second for loop, we iterate over all trips from $M$. For each trip $Tr_{i}$, we initialize an origin set $O$, and get origin POI $o_{i}$, destination POI $d_{i}$, and arriving time $d_{t_{i}}$ of the trip. After that, we traverse each trip $Tr_{j}$ from $M$, but except $Tr_{i}$ (Line 6), then we get origin POI $ o_{j} $, destination POI $d_{j}$, and the time of arriving to destination, which is $d_{t_{j}}$. If $d_{j}=d_{i}$ and $ |d_{t_{i}}-d_{t_{j}}| $ within the predefined time interval $ W_{d} $, we put $o_{j}$ in set $O$. Else, continue to check next trip. Finally, for origin POI $o_{i}$ of each trip, we generate an origin set $O$, and use it to update the representations (Line 13). We also use the Skip-gram to update these representations in accordance with our objective function in equation \ref{eq:od_model_log}.
 
In both Trip and OD models, our goal is to learn the matrix of place representations: $ \Phi \in \mathbb{R}^{|N|\times d} $, where $d$ is a small number of latent dimensions. The embedding represents place distribution context, and two places are similar when having trips to similar destinations.  We will analyze factors that will influence the performance of the two embedding models and use real datasets to compare and validate the two different models with baselines.
\section{Experiments}
The parameters we use for the experiments are shown in Table \ref{tab:parameters} with the bold values being the default setting.

\begin{table}[!ht]
\small
\centering
\caption{Parameters for Algorithms Comparison}
\begin{tabular}{|c|c|} \hline
\textbf{Parameter}&\textbf{Tested settings}\\ \hline \hline
\texttt{\textbf{Time (hours)}} & 0.5, \textbf{1}, 1.5, 2, 2.5, 3, 3.5, 4, 4.5, 5 \\ 
\hline
\texttt{\textbf{Dimension}}& 100, 110, 120, 130, 140, 150, 160, 170, \textbf{180}, 190, 200 \\ 

 \hline
\texttt{\textbf{Epoch}}& 2, 3, 4, 5, \textbf{6}, 7, 8, 9, 10  \\ 
\hline

\end{tabular}
\label{tab:parameters}
\end{table} 

\subsection{Dataset Description \& Preprocessing}
\subsubsection{New York City Taxi Dataset}
New York City Taxi \& Limousine Commission \cite{newyork} provides open access to yellow and green taxi trip records between 2009-2017. The green taxi dataset does not contain coordinates information, which we need to use in our experiment. So, in our work, we only focus on yellow taxi dataset. Table \ref{tab:trips} shows trips' information of yellow taxi. New York City Taxi \& Limousine Commission did not publish all the 2017 records, and there is no latitude and longitude information after June 2017. So we use the year 2015 taxi dataset as our experiment dataset, which contains 146,113,001 trips. For each trip record from yellow taxi dataset, it includes pick-up/drop-off times, the number of passengers, pick-up/drop-off locations represented by longitude and latitude, trip distance, rate types, payment type, and itemized fares.

\begin{table}[!ht]
\centering
\caption{New York City Yellow Taxi Dataset}
\begin{tabular}{|c|c|} \hline
\textbf{Year}&\textbf{Trips}\\ \hline \hline
\texttt{2016}& 131,165,055 \\ 
 \hline
\texttt{\textbf{2015}}& \textbf{146,113,001} \\ 
\hline
\texttt{2014}& 165,114,385 \\ 
\hline
\texttt{2013}& 173,179,783 \\ 
\hline
\texttt{2012}& 178,544,348 \\ 
\hline
\texttt{2011}& 176,897,223 \\ 
\hline
\texttt{2010}& 169,001,186 \\ 
\hline
\texttt{2009}& 170,896,079 \\ 
\hline
\end{tabular}
\label{tab:trips}
\end{table} 

\subsubsection{New York City POI Dataset}
\begin{figure}[ht]
        \centering
        \includegraphics[width=0.9\linewidth]{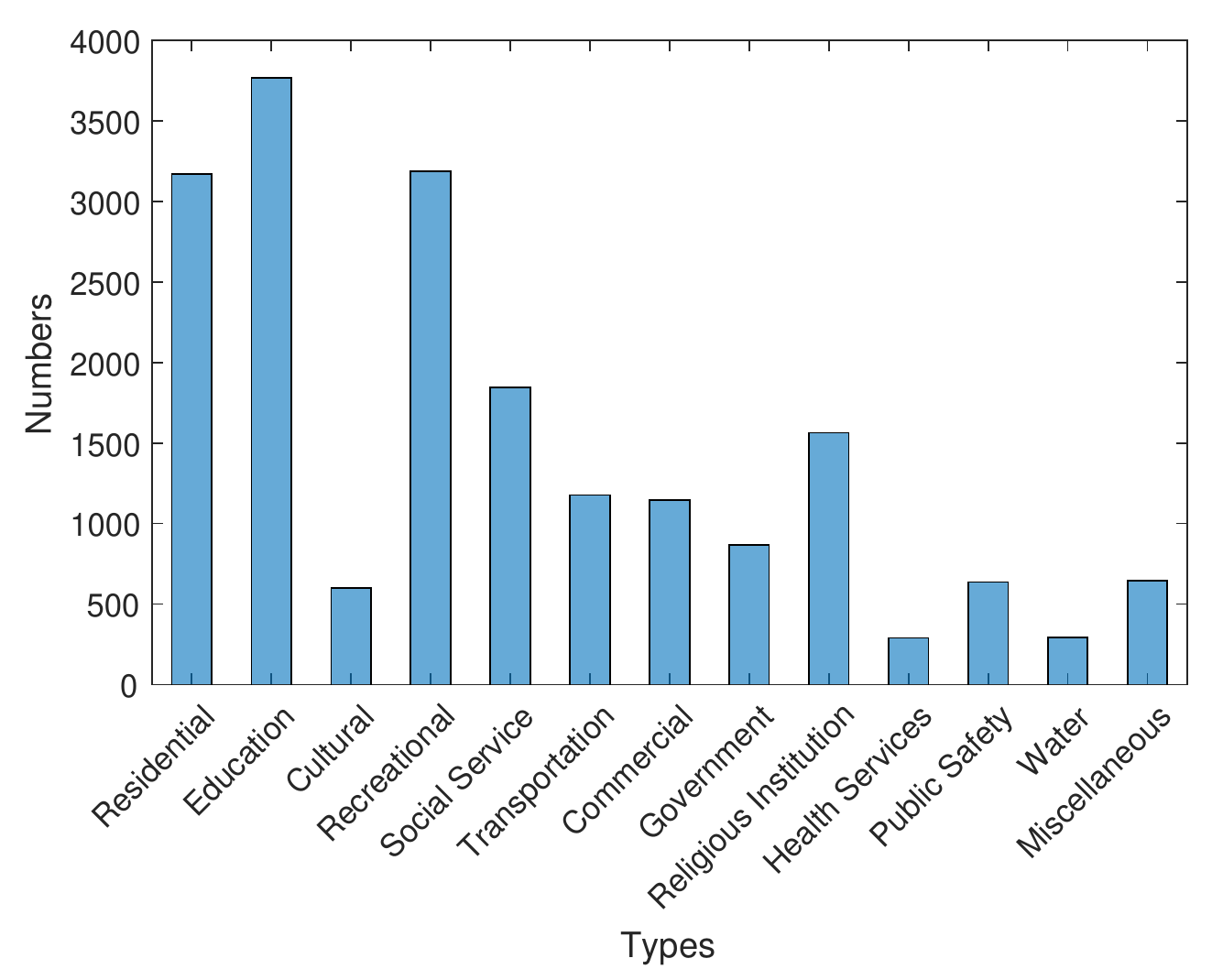}
        \caption{The Distribution of 13 POI Types}
        \label{fig:type}
\end{figure}

We collected the New York City POI data from NYC OpenData. There are 19,194 different POIs in total.  All the POIs are grouped into 13 different types. The distribution of 13 types are shown in figure \ref{fig:type}. For each POI, it contains coordinates (latitude and longitude), POI's unique ID, POI's address name and so on. 

\subsubsection{Dataset Preprocessing}  
In our New York City taxi dataset, there are some records missing coordinates information, so we need to remove those records. Moreover, in our problem, we need to assign the nearest POI to origin/destination point as the POI. We use 200 meters as range, and if there is no POI within 200 meters for origin/destination point, we remove the trip from the dataset. After these two preprocessing steps, we have 140,494,036 valid trips in total, which accounts for 98.2\% of original trips.

\subsection{Evaluation Metrics}
We define the similarity $\textit{S}$ of two POIs (\textit{i} and \textit{j}) using cosine similarity \cite{pennington2014glove} of the vector representations (\boldsymbol{$v_{i}$},\boldsymbol{$ v_{j}$}) of the POIs: 
\begin{equation}
S(\boldsymbol{v_{i}},\boldsymbol{ v_{j}})=cos(\boldsymbol{v_{i}},\boldsymbol{ v_{j}})=\dfrac{\boldsymbol{v_{i}}\cdot\boldsymbol{ v_{j}}}{\lVert\boldsymbol{v_{i}}\rVert\cdot\lVert\boldsymbol{ v_{j}}\rVert}
\end{equation}

The distances between two POIs is defined as \textit{d}(\boldsymbol{$v_{i}$},\boldsymbol{$ v_{j}$}) \cite{altszyler2016comparative}:
\begin{equation}
d(\boldsymbol{v_{i}},\boldsymbol{ v_{j}}) = 1-S(\boldsymbol{v_{i}},\boldsymbol{ v_{j}})
\end{equation}
 
\subsubsection{Match Rates}
For each POI $ i $ from test set, we use cosine similarity to find the most similar POI $ j $. We check the type of POI $ j $, if POI $j$ has the same type as POI $i$, there is a match and POI $ i $ is a matched  POI.  The match rate of a test set is the ratio of the matched POI in the set over the size of the test set as shown in equation \ref{match_rate}.

\begin{equation}
\label{match_rate}
match\_rate = \frac{number\ of\ matched\ POI}{number\ of\ POI\ in\ test\ set}
\end{equation}

\subsubsection{Silhouette Coefficients}
POIs of the same type form a cluster. Silhouette Coefficients measure how close the POI $ i $ is to other POIs within the same cluster compared to POIs in the closest cluster. We use Silhouette Coefficients as the cluster validity measure \cite{rousseeuw1987silhouettes}. The Silhouette Coefficients $ s(i) $ is defined as:

\begin{equation} \label{silhouette}
s(i) = \frac{b(i)-a(i)}{max\{a(i),b(i)\}}
\end{equation}
where $ a(i) $ is the mean distance of POI $ i $ to all other POIs within the same cluster, and $ b(i) $ is the minimum mean distance of POI $ i $ to all other POIs in another cluster. We use the mean value of all Silhouette Coefficients as the final Silhouette Score. The value of the score is between -1 to 1, the larger the value the tighter of the clusters: a $ s(i) $  value close to 1 means $ a(i)\ll b(i) $. Because $ a(i) $ is a distance measure and measures how dissimilar $ i $ to POIs of its own type, a small value of $ a(i) $ means $i$ is very similar to the POIs in its own cluster and a large value of $ b(i) $ means $ i $ is very different from POIs from other clusters. On the other hand, if $ s(i) $ close to -1, it means $ a(i)\gg b(i) $ and $ i $ has been assigned to the wrong cluster.

\subsection{Baselines}
Yan et al.\cite{yan2017itdl} defined four spatial context models based on the co-occurrence of center place type $ t_{center} $ and context place type $ t_{context} $, that incorporate distance decay and/or aggregated check-in data. The basic method of the spatial context models is the original Word2Vec. The key different between the four models is the definition of augmenting factor $ \beta  (  \beta  \in  \mathbb{Z} $ and  $ \beta \geqslant $1), which is used to increase the number of times of a tuple ($ t_{center} $, $ t_{context} $) appeared in the training dataset.

\textit{1) Check-in Context:} This context model incorporates human activities and defines $ P_{j} $ as the total number of check-ins of POI $ j $. In our case, $ P_{j} $ is the drop off numbers associated with POI $ j $. So the factor $ \beta $ is defined as:

\begin{equation}
\beta_{checkin}^{j} = \lceil 1+ln(1+P_{j})\rceil
\end{equation}
where $P_{j}$ is the context POI. 

\textit{2) Distance Context:} To incorporate distance information in context model, the augmenting factor $ \beta $ is defined as:

\begin{equation}
\beta_{distance}^{j} = \Bigg\lceil\frac{1+\frac{\sum_{k=1}^{\lvert L \rvert}P_{k}}{\lvert L \rvert}}{1+d^{\alpha}(i,j)}\Bigg\rceil
\end{equation}
in the equation, $ \lvert L \rvert $ is the total number of POIs, and $ d(i,j) $ is the distance between center POI $ i $ and context POI $ j $, $ \alpha $ is defined as inverse distance factor. In our experiment, we choose the same value of $ \alpha $ as Yan's work, which is set to 1.

\textit{3) Combined Context:} This context model uses both check-ins and distance information to decide the augmenting factor $ \beta $, which is defined as: 

\begin{equation}
\beta_{combined}^{j} = \Bigg\lceil\frac{1+ln(1+P_{j})}{1+d^{\alpha}(i,j)}\Bigg\rceil
\end{equation}

\textit{4) ITDL Context:} To get the best Information Theoretic Distance Lagged (ITDL) augmentation, it needs to find the relative importance between $ A $(activity) and $ U $(uniqueness) defined in equation \ref{eq:activity} and \ref{eq:uniqueness}. Same with Yan's work, we get the best $ \omega $ value (0.4) by testing the model with $ \omega $ value range from 0.1 to 1 with 0.1 as increment step. So the final augmenting factor $ \beta $ is given by equation \ref{eq:ITDL}.

\begin{equation} \label{eq:activity}
 A = -log_{2}\Bigg(1-\frac{P_{t_{j}}}{1+\sum_{k=1}^{\lvert M \rvert}P^{h}_{t_{k}}}\Bigg)  
\end{equation}
where $ t_{j} $ is the type of POI $ j $. So in equation \ref{eq:activity}, $P_{t_{j}} $ is the check-in numbers of POI type $ t_{j} $, $ \lvert M \rvert $ is the total number of types, and $ \sum_{k=1}^{\lvert M \rvert}P^{h}_{t_{k}} $ is total check-ins of all types within a distance bin, with width $ h $ (in the experiments $ h $ is set to 30 meters).
\begin{equation} \label{eq:uniqueness}
 U = -log_{2}(F^{h}_{t_{j}})  
\end{equation}
where $ F^{h}_{t_{j}} $ indicates the probability to encounter context place type $ t_{j} $ in distance bin $ h $.

\begin{equation} \label{eq:ITDL}
 \beta_{ITDL}^{j} = \lceil\omega A+(1-\omega)U\rceil  
\end{equation}

\subsection{Experiment Results}
\subsubsection{Performance w.r.t Dimension}
\begin{figure}[ht]
        \centering
        \includegraphics[width=0.9\linewidth]{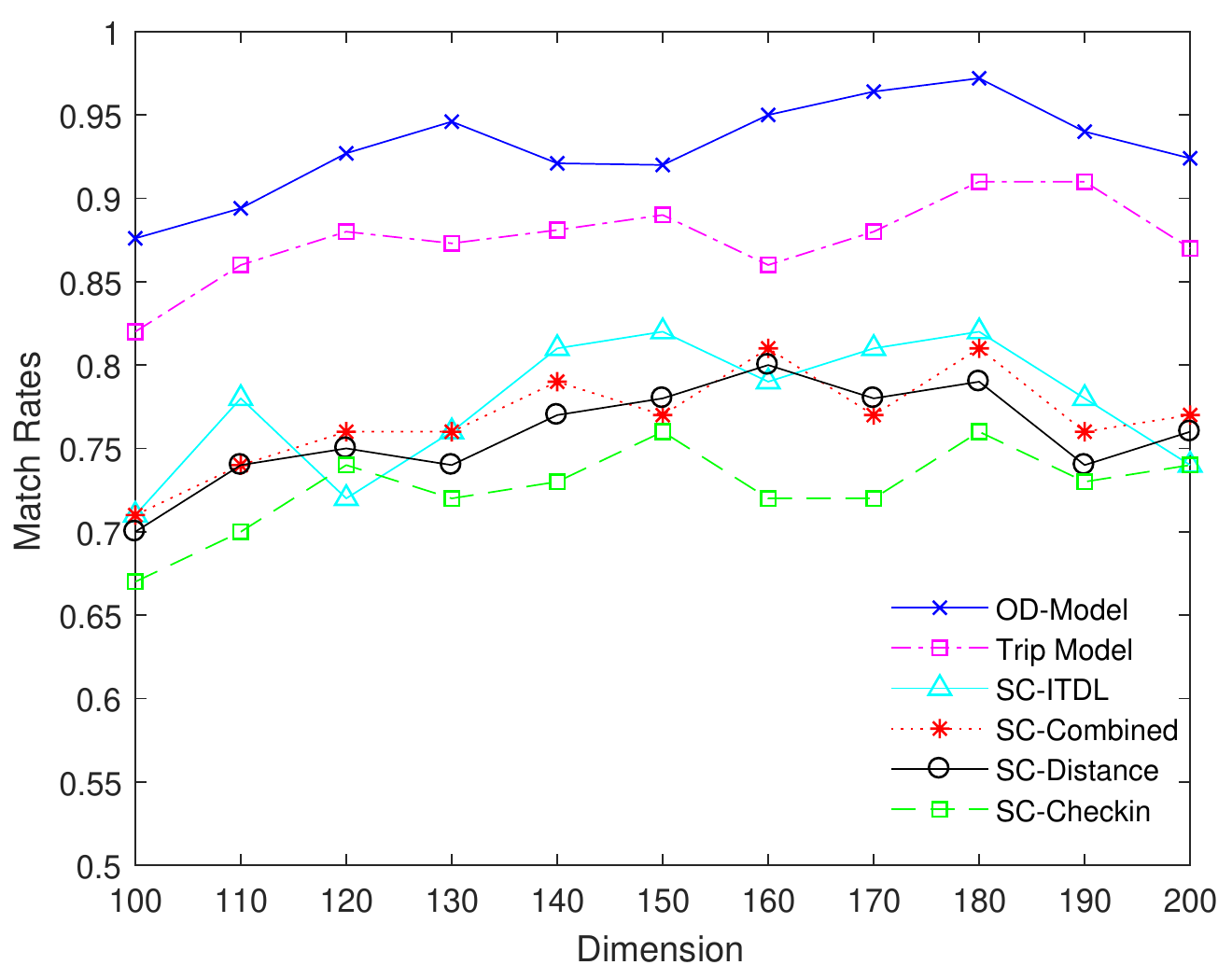}
       \caption{Match Rate of Six Models w.r.t Different Dimension Sizes}
       \label{fig:dimension}
\end{figure}

The dimension, also called number of features of the embedding vectors, is an important factor that influences the performance of latent representation models. Google's pre-trained Word2Vec model used 300 features as their embedding size. The model was trained on a vocabulary of 3 million words and phrases from Google News dataset. Compared with Google News dataset, we have a smaller dataset of 19,194 different POIs and 140 million trips. So we decide to compare the performance on dimensions ranging from 100 to 200 with 10 as incremental step size. Figure \ref{fig:dimension} shows the match rate with different dimensions. From the Figure \ref{fig:dimension} we can see that both of the Trip model and OD model outperform the four baselines significantly. When dimension is 180, all models achieve the best results. The best match rate of four baselines is 0.82, which is achieved by using ITDL model. The Trip model and OD model have match rates at 0.91 and 0.97 respectively. OD model has a better performance than all other models including Trip model for all dimensions tested.

\begin{figure}[ht]
        \centering
        \includegraphics[width=0.9\linewidth]{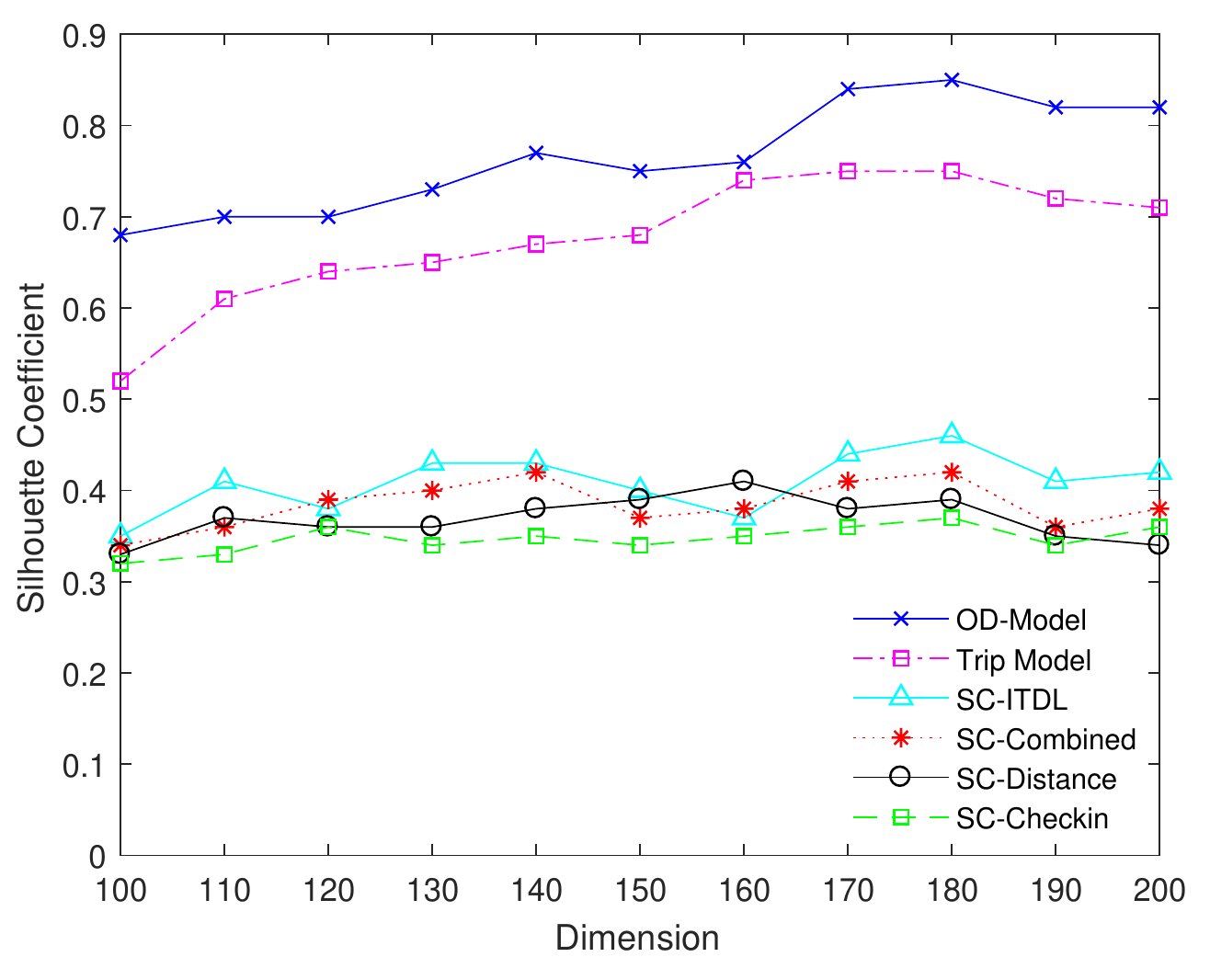}
      \caption{Silhouette Coefficients of Six Models w.r.t Different Dimension Sizes}
      \label{fig:sc_dimension}
\end{figure}

Figure \ref{fig:sc_dimension} shows the Silhouette Coefficient scores of six methods with different dimensions. We can find that Trip model, and OD model have a much higher score than four baselines, which indicates our proposed Trip model and OD model could cluster the data appropriately. The lowest Silhouette Coefficient score of Trip model and OD model is 0.52, which is still better than the highest score 0.46 of four baselines. Compared with the Trip model, the OD model performs better, which can achieve the highest Silhouette Coefficient score. OD model is consistently 30\% to 39\% better than ITDL model for all dimensions. 

\subsubsection{Performance w.r.t Time Interval}
\begin{figure}[ht]
        \centering
        \includegraphics[width=0.9\linewidth]{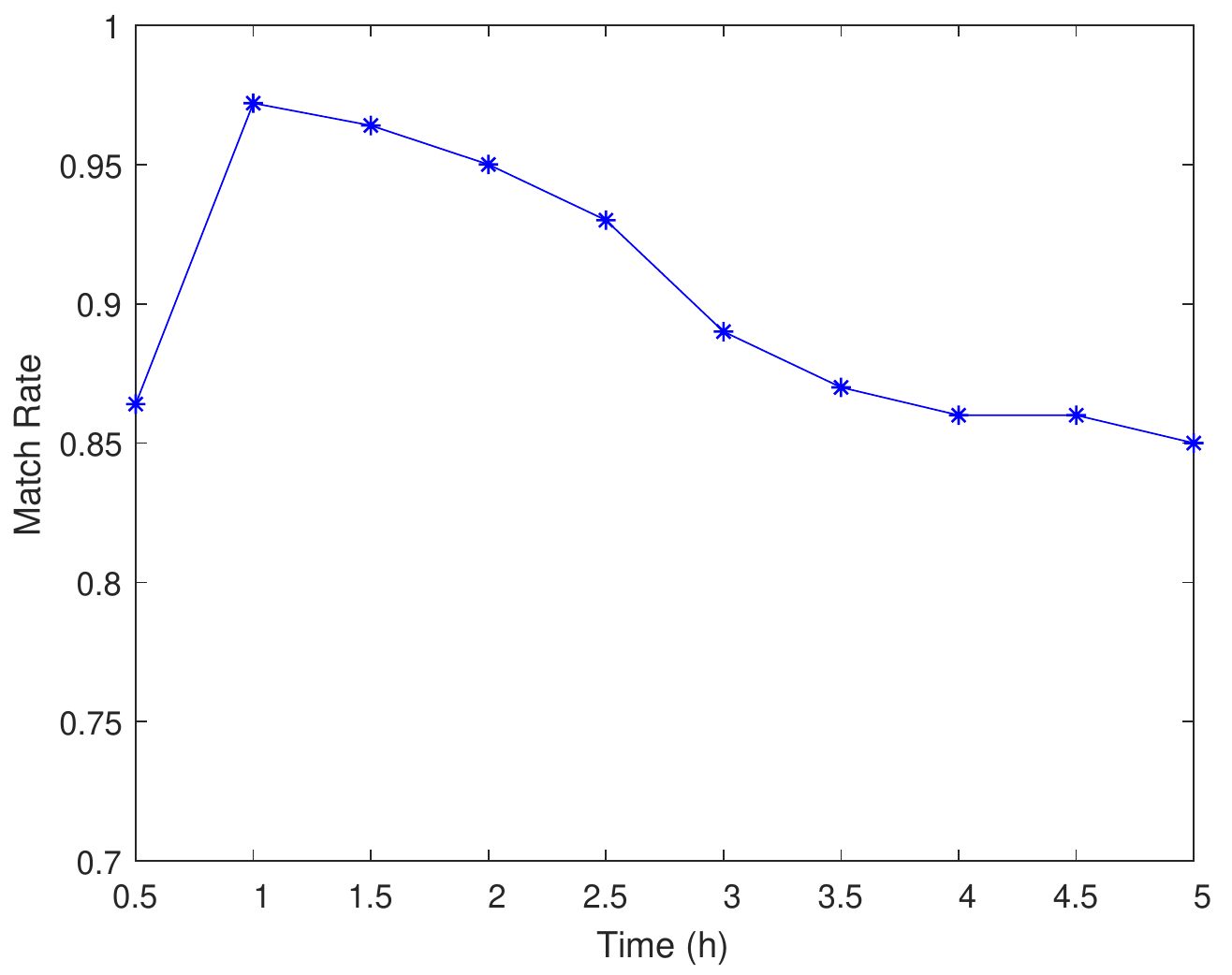}
      \caption{Match Rate of OD Model w.r.t Different Time}
      \label{fig:time}
\end{figure}

Temporal information of movements is a crucial factor for place representations. In our OD model, we incorporate time interval $W_d$ when building the context model. So, time interval will influence the performance of our OD model. Figure \ref{fig:time} shows the match rate with different time intervals. In our work, we start the time interview from 0.5 hours and end with 5 hours, and the incremental step is 0.5 hour. From the figure, we can see that, as the time interval increases from 0.5 hours to 1 hour, the match rate increase rapidly, which means the in 1 hour time interval, it contains more proper movements than 0.5 hours. When the time interval greater than 2.5 hours, the curve falls fast because the time interval includes many irrelevant movements. The best match rate is achieved at the time interval of 1 hour with 0.97 as the match rate, and the lowest match rate is 0.85 at the time interval of 5 hours.

\begin{figure}[ht]
        \centering
        \includegraphics[width=0.9\linewidth]{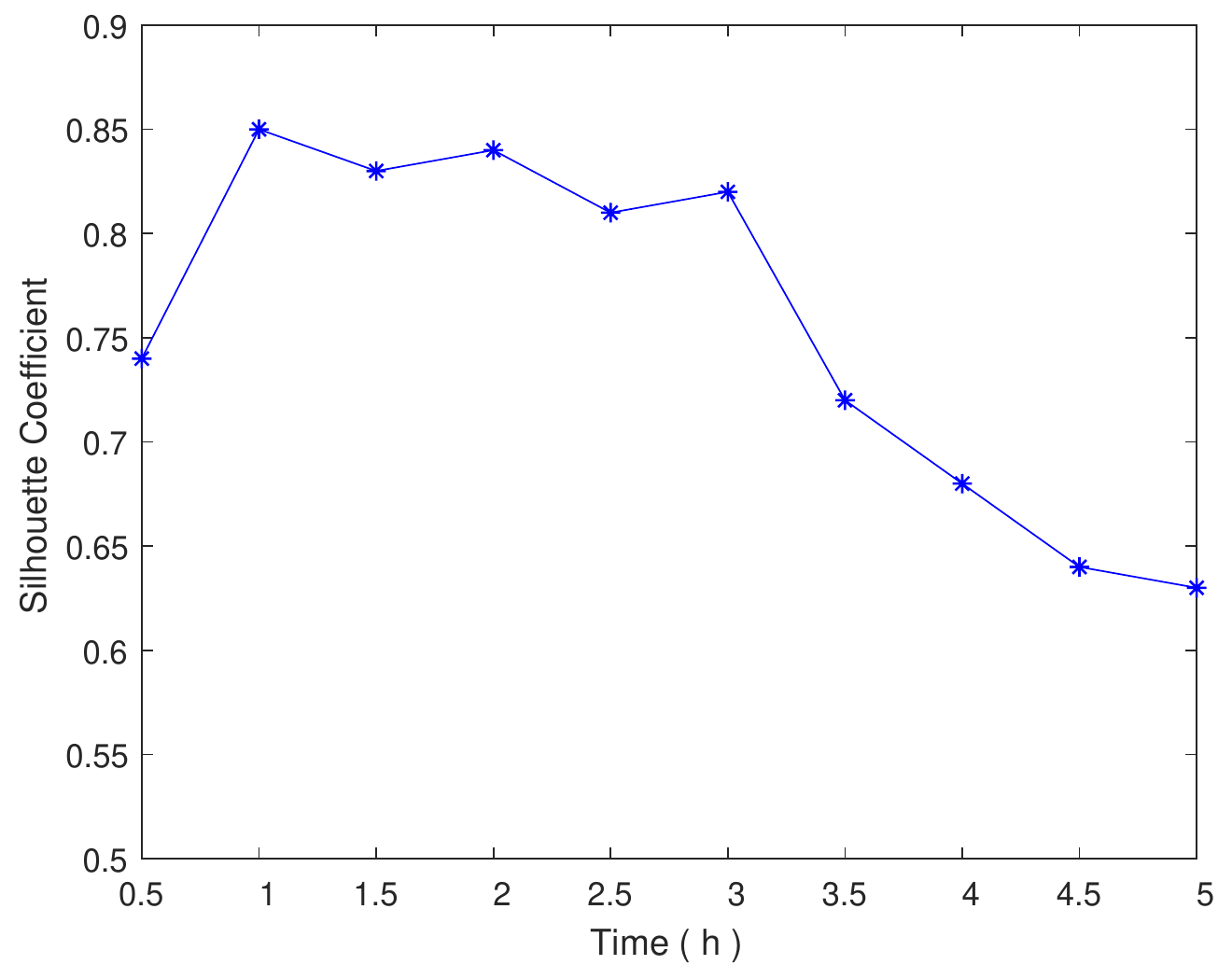}
      \caption{Silhouette Coefficients of OD Model w.r.t Different Time}
      \label{fig:time_sc}
\end{figure}

Figure \ref{fig:time_sc} shows Silhouette Coefficient score with different time intervals, and we can also find that using 1 hour as the time interval, OD model can achieve the highest scores of 0.85. It means with 1 hour as the time interval, and we can get the proper number of relative movements.  

\subsubsection{Performance w.r.t Epoch}
\begin{figure}[ht]
        \centering
        \includegraphics[width=0.9\linewidth]{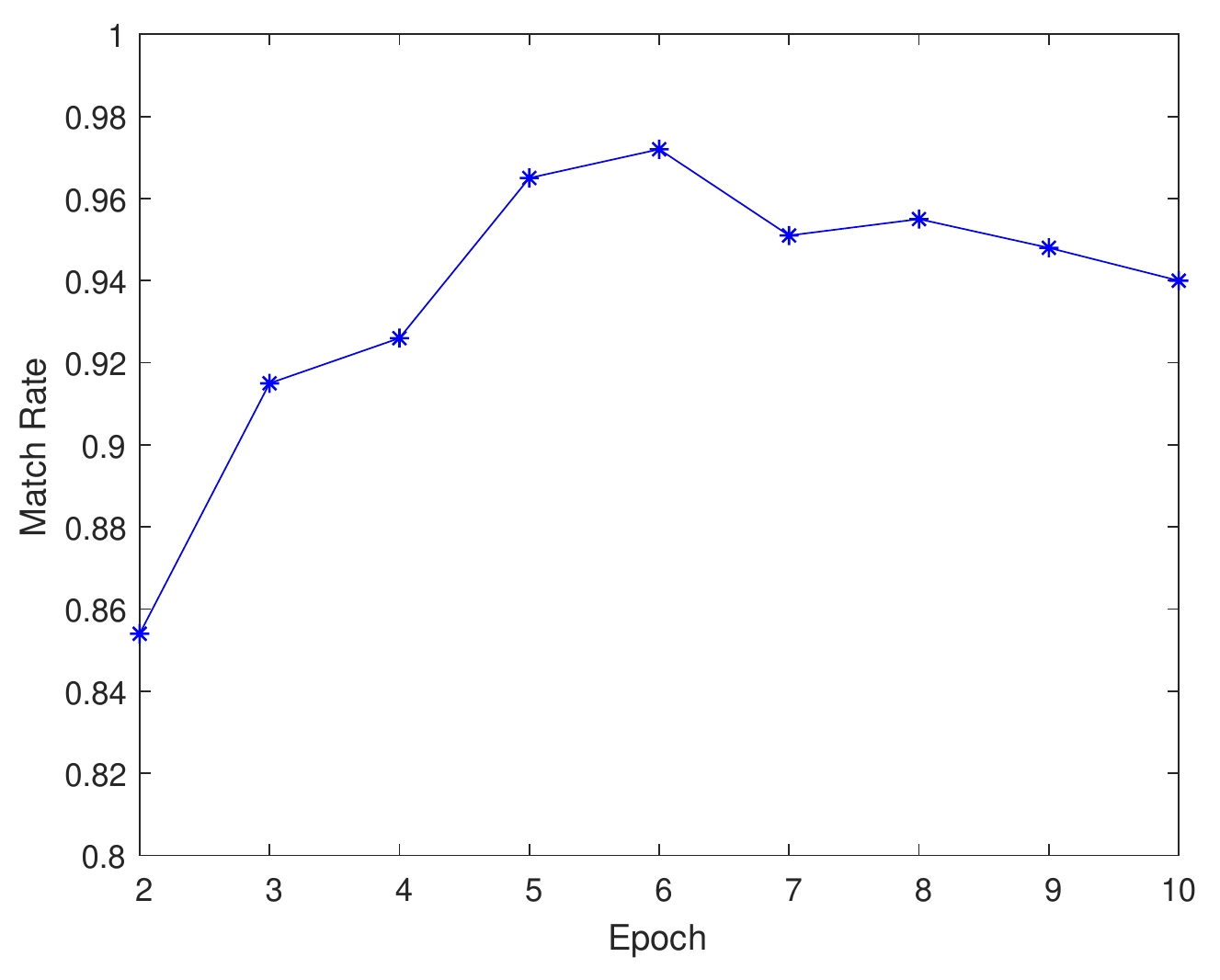}
      \caption{Match Rate of OD Model w.r.t Different Epoch}
      \label{fig:epoch}
\end{figure}

An epoch is a single pass through the full training dataset. Our model is trained until the error rate is acceptable, and this will often take multiple passes through the complete dataset. However, a large number of epochs not only takes very long training time but also will cause overfit. So, the number of epochs is another factor, which can influence the performance of our OD model. Figure \ref{fig:epoch} shows the experiment results of match rate under the different number of epochs. From the Figure we can see that, as epoch number increases from 2, the match rate also increases. It means that, when epoch number is 2, the match rate of OD model is only 0.854, which means the OD model could not get all the information from the dataset, and as we increase epoch number, the match rate increases and the model can learn more from the dataset. When the epoch number reaches 6, our model can achieve the best match rate. After that, the curve declines due to the overfitting.

\begin{figure}[ht]
        \centering
        \includegraphics[width=0.9\linewidth]{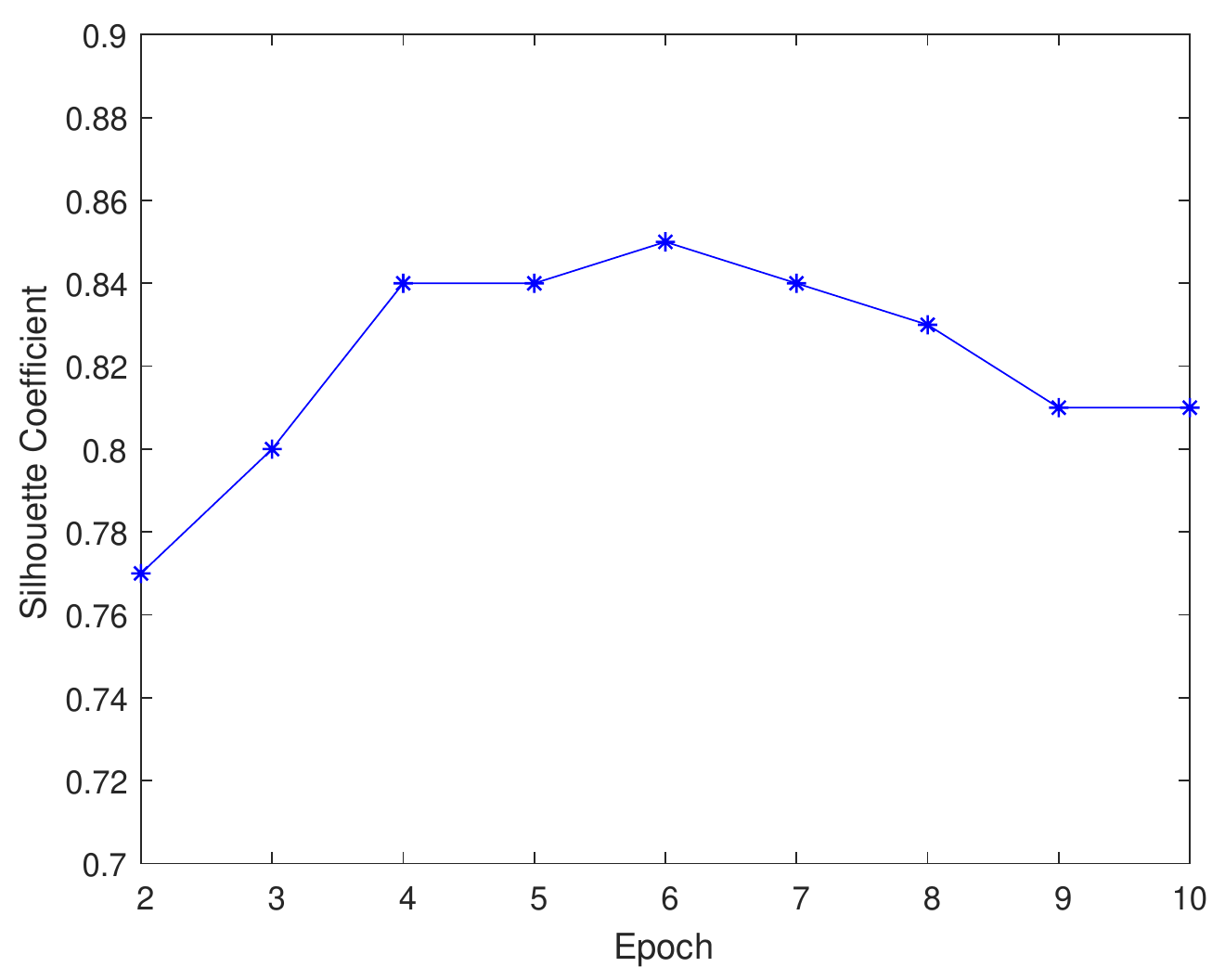}
      \caption{Silhouette Coefficients of OD Model w.r.t Different Epoch}
      \label{fig:sc_epoch}
\end{figure}

Figure \ref{fig:sc_epoch} shows the relation between Silhouette Coefficients and number of epochs, which gives the view of the performance with respect to epoch. The results from figure show that the highest Silhouette Coefficient score is 0.85 and achieved with 6 epochs, after that, it decreases to 0.81.  

\subsubsection{Performance w.r.t Data Size}
\begin{figure}[ht]
        \centering
        \includegraphics[width=0.9\linewidth]{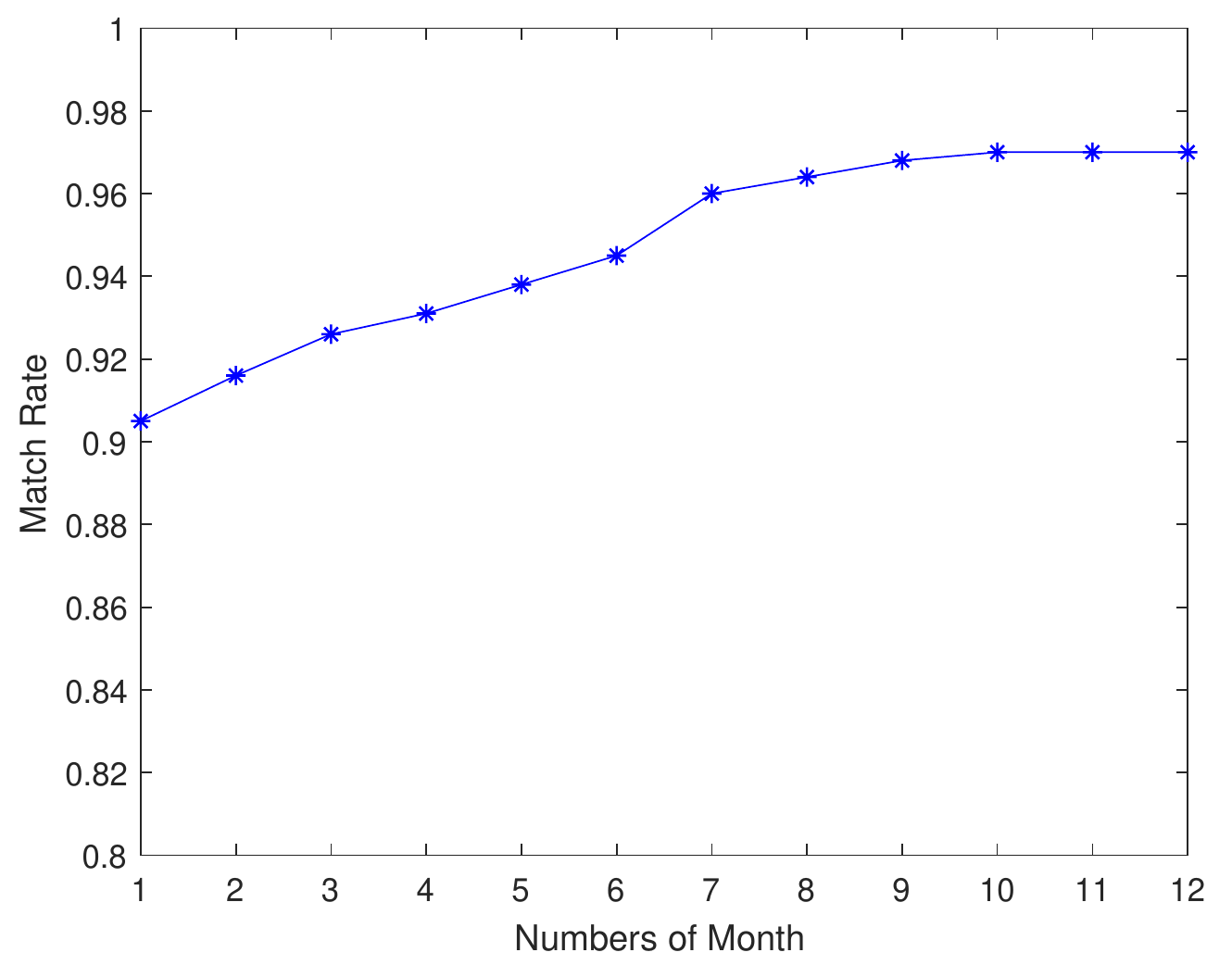}
      \caption{Match Rate of OD Model w.r.t Different Data Size}
      \label{fig:datasize}
\end{figure}

\begin{figure}[ht]
        \centering
        \includegraphics[width=0.9\linewidth]{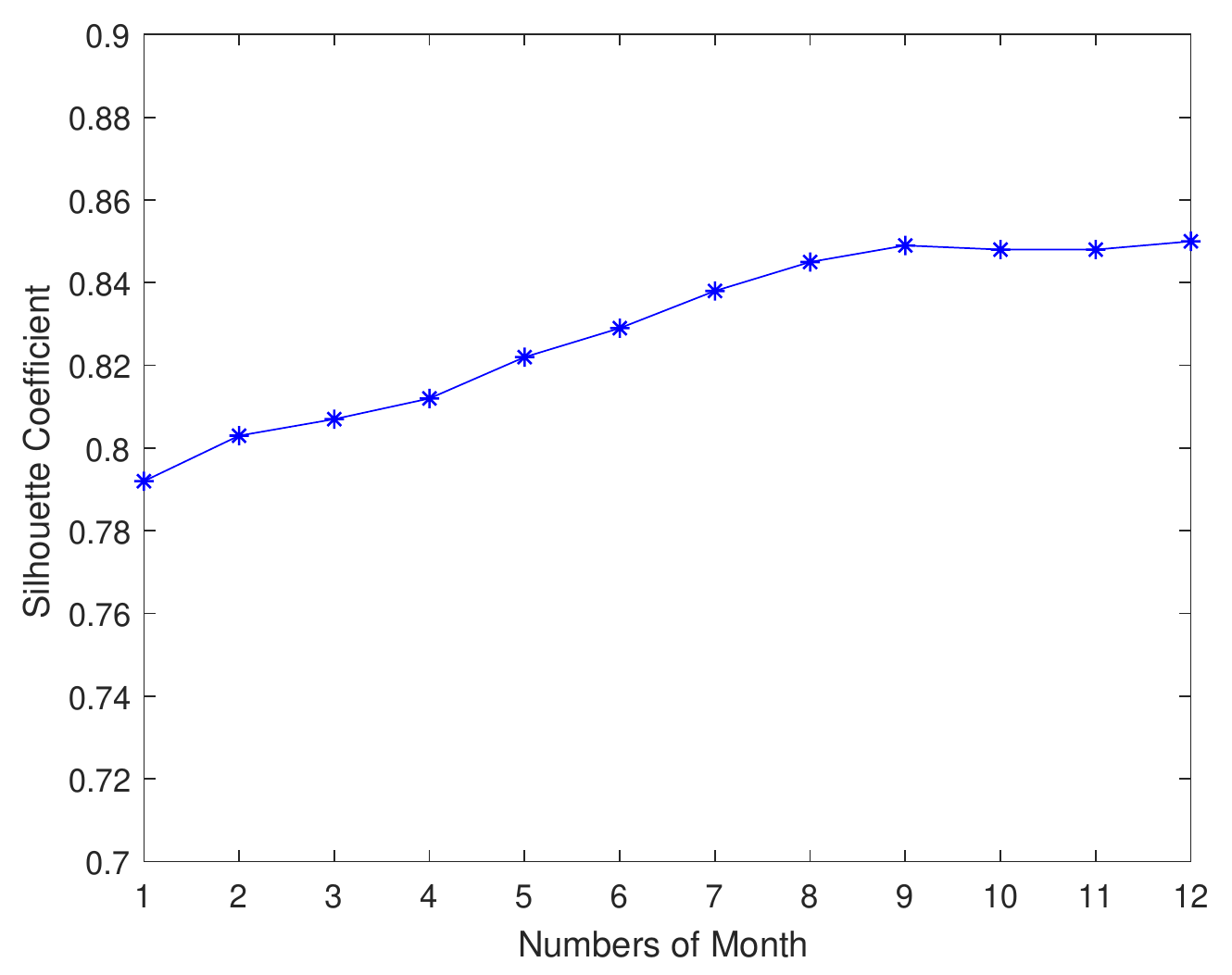}
     \caption{Silhouette Coefficients of OD Model w.r.t Different Data Size}
     \label{fig:sc_datasize}
\end{figure}

The size of the dataset is also an important factor. To get a good model, we need a certain number of the dataset to make sure the model can learn enough information from the data. So, we need to determine the optimum size of the training dataset necessary to achieve high classification match rate or Silhouette Coefficients score. In our experiment, we select the dataset ranging from a month to a year with increasing step as the one-month to decide the optimum size of the dataset. We can see the result from Figure \ref{fig:datasize} which shows that the match rate of 1 month dataset is 0.904. As the dataset increases, the match rate also increases until the dataset is enlarged to 9 months where the match rate arrives its highest match rate.
After that, the curve reaches a steady state and does not change much in match rate regardless of data size. We find the relationship between Silhouette Coefficient score and data size from Figure \ref{fig:sc_datasize}. 

We can see that the Silhouette Coefficients score increases with increased dataset until it reaches the certain size of the dataset, then it becomes stable. So, the number of datasets indeed has influences on our model. 
Figure \ref{fig:datasize} also shows that OD model trained on 1 month data achieves 90\% matching rate and 0.79\% Silhouette Coefficient value which is better than ITDL model trained on 12 months of data with 80\% matching rate and 0.45\% Silhouette Coefficient value.

\section{Related Work}
\subsection{Point-of-interest embeddings}

Xie et al. \cite{Xie:2016:LGP:2983323.2983711} develop a graph-based embedding (GE) model to capture the sequential effect, geographical influence, temporal cyclic effect and semantic effect in a unified way by embedding the four corresponding relational graphs (POI-POI, POI-Region, POI-Time, POI- Word) into a shared low dimensional space for the ease of data sparsity, cold-start problem, and context-aware recommendation in LBSNs.

Yan et al. \cite{yan2017itdl} proposed a model to capture the semantics information of place types. The model is based on Word2Vec, which augments the spatial contexts of POI types by using distance and information-theoretic approach to generate embedding. Then they used the generated vector embeddings to reason about the similarity and relatedness among place type. However, in their method, they did not consider movement information, which is important information when analyzing POI embeddings.  Movement of people can reflect the relationship among the POIs. In Yan's work, they consider distance, and check-ins information in their model, which sometimes could not reflect the true information of the similarity among POIs. 

Feng et al. \cite{feng2017poi2vec} propose a POI2Vec latent representation model, which uses the geographical influence of POIs to learn latent representations. Based on POI2Vec model, they further develop a method to model POI sequence and user preference. By using check-ins data, they try to estimate the probability of visiting a POI given previous visited POIs. However, compared with trip dataset, check-ins data has some limitations: data is small and  incomplete. For example, some people may not check in all the time, which causes the data not reflecting the movement information. When incorporating geographical information, they make a simple assumption and only consider nearby POIs that has high relevance. The model does not consider distribution of movement origins and destinations. 

Pang et al. \cite{pang2016deepcity} introduce a model called DeepCity, which is based on deep learning. It is used to learn features for user and location profiling. They propose a task-specific random walk method, which uses location and user information to guide learning algorithms to embed locations with similar categories. They collect the check-in data from Instagram. However, Instagram has many private users. Data privacy is a concern and it is difficult to get enough data for a city. Our proposed model is more general and not task specific.

\subsection{Network embeddings}

Social network representations learning is getting more and more attention \cite{perozzi_deepwalk:_2014,tang_line:_2015,cao_grarep:_2015,grover_node2vec:_2016,wang_structural_2016,wang_community_2017}. Perozzi et al. \cite{perozzi_deepwalk:_2014} propose DeepWalk to learn latent representations of vertices in a unweighted network using local information from truncated random walks as input. The skip-gram model with hierarchical softmax is used as the loss function. Tang et al. \cite{tang_line:_2015} propose LINE which works with undirected, directed, and/or weighted networks. They use edge-sampling for inference and try to preserve both the local and global network structures by using first-order and second-order proximities as objective functions, an approach Wang et al. \cite{wang_structural_2016} also exploit later. Cao et al. \cite{cao_grarep:_2015} integrate global structural information of the graph into the learning process by using k-step loss functions. Grover et al. \cite{grover_node2vec:_2016} argue that the added flexibility in exploring neighborhoods is the key to learning richer representations. They propose node2vec which defines a flexible notion of a node’s network neighborhood and design a biased random walk procedure, which explores diverse neighborhoods. In addition to microscopic structure such as the first- and second-order proximities, Wang et al. \cite{wang_community_2017} also try to preserve community structure of the network by using a matrix factorization model.

Yang et al. \cite{yang_revisiting_2016} argue that the graph embeddings \cite{perozzi_deepwalk:_2014,tang_line:_2015} are usually learned separately from the supervised task and do not leverage the label information in a specific task; therefore, they do not produce useful features and might not be able to fully leverage the distributional information encoded in the graph structure. In their semi-supervised learning framework, the embedding of an instance is jointly trained to predict the class label of the instance and the context in the graph. In another effort to employ nodes' information, Huang et al. \cite{huang_label_2017} propose a Label informed Attributed Network Embedding (LANE) framework which incorporates label information into the attributed network embedding while preserving their correlations.

While the works above study homogeneous networks in which there is a single type of nodes and links, Dong et al. \cite{dong_metapath2vec:_2017} consider heterogeneous networks with multiple types of nodes such as authors, papers, venues, etc. They develop a meta-path-guided random walk strategy to capture the structural and semantic correlations of differently typed nodes and relations.

\subsection{Word embeddings}
Distributed representation is a way to represent data using the structure of artificial neural networks (abbreviated as ANNs or NNs). Rumelhart et al. \cite{rumelhart_learning_1986} show that back-propagation for ANNs can be used to construct such representations of the inputs. Years later, Bengio et al. \cite{bengio_neural_2003} propose a model which learns simultaneously a distributed representation for each word along with the probability function for word sequences (language models), expressed in terms of these representations. Their language models outperform state-of-the-art n-gram models at the time. In \cite{collobert_unified_2008}, Collobert and Weston propose a general deep NN architecture for natural language processing (NLP), which is jointly trained on multiple tasks such as semantic role labeling, named entity recognition, part-of-speech tagging, chunking, and language modeling. Their model uses a lookup-table layer to map words into real-valued vectors.

Reisinger and Mooney \cite{reisinger_multi-prototype_2010} argue that capturing the semantics of a word with a single vector is problematic. They then present a method that uses clustering to produce multiple “sense-specific” vectors for each word, accommodating homonymy and polysemy. Huang et al. \cite{huang_improving_2012} propose a NN architecture which learns word embeddings that capture the semantics of words by incorporating both local and global document context, and accounts for homonymy and polysemy by learning multiple embeddings per word.

Mikolov et al. \cite{mikolov_distributed_2013, mikolov_efficient_2013, mikolov_linguistic_2013} show that it is possible to train quality word vectors using simple model architectures (word2vec), compared to the popular neural network models (both feedforward and recurrent). The lower computational complexity enables accurate high dimensional word vectors from large datasets. They also demonstrate that these representations exhibit linear structure that makes precise analogical reasoning possible, and word vectors can be somewhat meaningfully combined using just simple vector addition. The learned word representations capture syntactic and semantic regularities which are observed as constant vector offsets between pairs of words sharing a particular relationship.

In \cite{pennington_glove:_2014}, Pennington et al. argue that count-based and prediction-based distributed word representations are not dramatically different since they both probe the underlying co-occurrence statistics of the corpus. They then construct a model (GloVe) that utilizes the global statistics benefit of count data while simultaneously capturing the meaningful linear substructures prevalent in log-bilinear prediction-based methods like word2vec. GloVe outperforms other models on word analogy, word similarity, and named entity recognition tasks. Levy et al. \cite{levy_improving_2015} show that much of the performance gains of word embeddings are due to certain system design choices and hyperparameter optimizations, rather than the embedding algorithms themselves. These modifications can be transferred to traditional count-based distributional models, yielding similar gains. They observe mostly local or insignificant performance differences between the methods, with no global advantage to any single approach over the others.

Besides learning word representations, there are efforts \cite{cho_learning_2014,le_distributed_2014,arora_simple_2016} addressing representations of phrases, sentences, and documents. These representations improve the performance of machine translation system, and achieve positive results on some text classification and sentiment analysis tasks. There are more and more applications of word embeddings in not only natural language tasks \cite{lample_neural_2016,cotterell_probabilistic_2017}, but also in computer vision, e.g., Karpathy and Fei-Fei \cite{karpathy_deep_2015} introduce a model that aligns parts of visual and language modalities to generate natural language descriptions of image regions based on weak labels in form of a dataset of images and sentences.

\section{Conclusion}
In this paper, we propose DeepMove, a novel approach to learning the latent representations of places. Using  movements information as input, our method learns a representation, which can incorporate the spatial and temporal semantics of places and be used for analyzing place similarity and relationships. Experiments on New York City taxi trip dataset illustrate the effectiveness of our approach on capturing the latent semantic information for places. We propose two different evaluation methods to evaluate place embeddings. We compare our methods with four baselines. Results show that our models outperform baselines. Moreover, the results also validate the importance of movements on place embedding. Finally, we give a detailed analysis of different factors, which will influence the performance of OD model to gain a better understanding of our work.

In the future, we will use the latent representation (embedding) from our current work with traditional features such as text description and check-ins from Yelp, Tweets and so on, to classify places, estimate flows, and detect anomaly places.

\bibliographystyle{ACM-Reference-Format}
\bibliography{sigproc} 

\thispagestyle{plain}

\end{document}